# A Novel Spatiotemporal Coupling Graph Convolutional Network

Fanghui Bi

*Abstract*—Dynamic Quality-of-Service (QoS) data capturing temporal variations in user-service interactions, are essential source for service selection and user behavior understanding. Approaches based on Latent Feature Analysis (LFA) have shown to be beneficial for discovering effective temporal patterns in QoS data. However, existing methods cannot well model the spatiality and temporality implied in dynamic interactions in a unified form, causing abundant accuracy loss for missing QoS estimation. To address the problem, this paper presents a novel Graph Convolutional Networks (GCNs)-based dynamic QoS estimator namely Spatiotemporal Coupling GCN (SCG) model with the three-fold ideas as below. First, SCG builds its dynamic graph convolution rules by incorporating generalized tensor product framework, for unified modeling of spatial and temporal patterns. Second, SCG combines the heterogeneous GCN layer with tensor factorization, for effective representation learning on bipartite user-service graphs. Third, it further simplifies the dynamic GCN structure to lower the training difficulties. Extensive experiments have been conducted on two large-scale widely-adopted QoS datasets describing throughput and response time. The results demonstrate that SCG realizes higher QoS estimation accuracy compared with the state-of-the-arts, illustrating it can learn powerful representations to users and cloud services.

## I. Introduction

In the contemporary web ecosystem, cloud services are fundamental, enabling interoperability among heterogeneous software applications. The rapid expansion of cloud computing, e-commerce, and the Internet of Things (IoT) has led to a proliferation of cloud services, resulting in a market oversaturation that complicates the selection process for users [1]-[3]. The evaluation of Quality-of-Service (QoS) data is paramount in this context, as it informs service selection based on attributes such as robustness, cost, and response time, which are assessed from both the user and server perspectives [4]-[7]. Recently, researchers have delved into QoS estimation methods to accurately predicting unknown QoS values based on historical records. However, QoS matrices are often High-Dimensional and Incomplete (HDI) due to the scarcity of achieved invocations compared to the total possible ones. In such scenario, Latent Feature Analysis (LFA) has garnered attention owing to its effectiveness in handling a HDI matrix [1], [5]-[9]. LFA maps users and services into a latent feature space, leveraging these features to estimate missing QoS data.

In practice, QoS values of cloud services often fluctuate over time as illustrated in Fig. 1, emphasizing the need to explore temporal patterns. Hence, several dynamic LFA-based QoS estimators are proposed to capturing latent features based on a QoS tensor [10]-[20]. While existing methods for dynamic QoS estimation have demonstrated effectiveness, they cannot well capture the spatiotemporal information in a uniform way. Specifically, they suffer from the following limitations: a) ***Incomplete Representation:*** They tend to separate the modeling of temporal and spatial patterns, which prevents unified handling of dynamic non-Euclidean structure and results in information loss, as shown in Fig. 1. b) ***Complex Learning Paradigm:*** For the unattributed graphs, they commonly construct attributes and apply complicated message propagation rules, which cannot learn expressive latent features [21]-[24].

With the above findings, we pose the following Research Question (RQ): *Is it possible to uniformly capture the spatiotemporal patterns implied in dynamic QoS data to perform effective LFA for accurately estimating the missing QoS values?* To answer it, this study develop a novel dynamic QoS estimator based on **S**patiotemporal **C**oupling **G**CNs, named as SCG. The model is achieved with the three-fold ideas: a) building a novel graph convolution rule based on generalized tensor product framework [25]-[26]; b) combing the dynamic heterogeneous GCN layer with tensor factorization for effective learning to latent features; and c) removing redundant message propagation operations to lower the training difficulties and improve learning efficiency. To be specific, this paper has the contributions as follows:

- We propose a novel dynamic GCN layer, capturing richer spatiotemporal information implied in QoS data, which can be extended to other downstream tasks.
- We present a SCG model used for temporal-aware QoS estimation, which incorporates the proposed GCNs into tensor factorization for accurate LFA.
- Comprehensive empirical studies are conducted on two large-scale industrial QoS datasets, and the comparison results with several state-of-the-art methods demonstrate it can perform more accurate estimation.

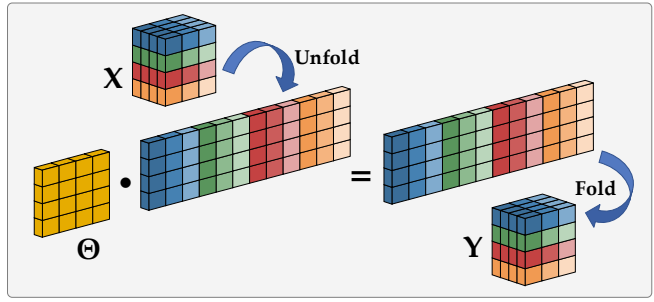

Figure 1. An illustrate example of Θ-transform on a 4×3×4 tensor.

The rest of this paper is well organized as below. Section II presents the preliminaries. Section III introduces the SCG model. Section IV analysis the experimental results. At last, Section V concludes this paper.

## II. PRELIMINARIES

To enhance the description of this paper, we adopt the following conventions: Three-dimensional tensors are denoted by uppercase bold letters, e.g., **X**; matrices are represented by uppercase non-bold letters, e.g., X; sets are indicated by using uppercase italic letters, e.g., *X*; and the elements within tensors, matrices, or sets are represented by lowercase italic letters, e.g., $x_{i,j,k}$.

## III. THE SCG MODEL

### A. Dynamic GCNs based on Generalized Tensor Product

The normal graph convolution process is defined as [13]-[20], [25]-[32]:

$$\begin{cases} X^{(l+1)} = \sigma\left(\hat{A} \cdot X^{(l)} \cdot W^{(l)}\right), \\ \hat{A} = \hat{D}^{-1/2} \cdot (A+I) \cdot \hat{D}^{-1/2}, \end{cases} \quad (1)$$

where $X^{(l)}$ is the *l*-th layer feature matrix; A is the adjacency matrix; *I* is the identity matrix; $\hat{D}$ is the diagonal degree matrix, whose each element $d_{i,i}=1+\sum_j a_{i,j}$; $W^{(l)}$ is the *l*-th layer feature transformation matrix; and $\sigma(\cdot)$ is the activation function, such as Sigmoid. Now, given adjacency tensor **A** and feature tensor **X**, we aim to build a spatiotemporal GCN layer base on them. To achieve this, based on the calculation rules in Fast Fourier Transform (FFT) [25]-[26], we define the generalized tensor product used in our study, which involves three fundamental definitions as follows.

***Definition 3-1. (Θ-Transform).*** Given a tensor $\mathbf{X}^{|I|\times|J|\times|T|}$ and a mixing matrix $\Theta^{|T|\times|T|}$, the Θ-transform of $\mathbf{X}^{|I|\times|J|\times|T|}$ is defined as $\mathbf{X}\otimes\Theta=\mathbf{Y}^{|I|\times|J|\times|T|}$, where each element is given as:

$$y_{i,j,t} = \sum_{r=1}^{|T|} \theta_{t,r} \cdot x_{i,j,r}. \quad (2)$$

***Definition 3-2. (Facewise Product).*** Two tensors $\mathbf{X}^{|I|\times|J|\times|T|}$ and $\mathbf{Y}^{|J|\times|K|\times|T|}$ are given, the facewise product between them is denoted as $\mathbf{X}\odot\mathbf{Y}=\mathbf{Z}^{|I|\times|K|\times|T|}$, whose element is calculated as:

$$z_{i,j,t} = \sum_{j=1}^{|J|} x_{i,j,t} \cdot y_{j,k,t}. \quad (3)$$

***Definition 3-3. (Θ-Product).*** Given two tensors $\mathbf{X}^{|I|\times|J|\times|T|}$ and $\mathbf{Y}^{|J|\times|K|\times|T|}$ and a mixing matrix $\Theta^{|T|\times|T|}$, the Θ-product between **X** and **Y** is calculated as $\mathbf{X}*\mathbf{Y}=(\mathbf{X}\otimes\Theta)\odot(\mathbf{Y}\otimes\Theta)=\mathbf{Z}^{|I|\times|K|\times|T|}$.

Based on the above definitions, a single spatiotemporal GCN layer is achieved as:

$$\begin{cases} \mathbf{X}^{(l+1)} = \sigma\left(\hat{\mathbf{A}} * \mathbf{X}^{(l)} * \mathbf{W}^{(l)}\right), \\ \hat{\mathbf{A}} = \hat{\mathbf{D}}^{-1/2} \odot (\mathbf{A}+\mathbf{I}) \odot \hat{\mathbf{D}}^{-1/2}, \end{cases} \quad (4)$$

where the tensors are the extensions of the matrices in (1).

### B. Layer-Wise Graph Convolution on a QoS Tensor

In the task of QoS estimation, (4) is extended to build a dynamic heterogeneous GCN layer for message propagation on the bipartite user-service interaction graphs. Specifically, the following computations are given as:

$$\begin{cases} \mathbf{U}^{(l+1)} = \sigma\left(\hat{\mathbf{A}} * \mathbf{S}^{(l)} * \mathbf{W}^{(l)}\right), \\ \mathbf{S}^{(l+1)} = \sigma\left(\hat{\mathbf{A}}^T * \mathbf{U}^{(l)} * \mathbf{B}^{(l)}\right). \end{cases} \quad (5)$$

where $\hat{\mathbf{A}}^T \in \mathbb{R}^{|U| \times R \times |T|}$ is the transposed tensor of $\mathbf{A}$ along the time dimension. It is worth noting that in (5), information flows between two distinct types of nodes: users and cloud services. This process captures high-order spatiotemporal connectivity through layer-by-layer computations.

Since the initial input node features for (5) are obtained using tensor factorization, the essential parameter tensors to be optimized consist of $\mathbf{U}^{(0)}$, $\mathbf{S}^{(0)}$, $\mathbf{W}^{(0 \sim L-1)}$, and $\mathbf{B}^{(0 \sim L-1)}$, where $L$ is the layer number. However, prior research on matrix factorization has indicated that coupling such parameters can lead to increased training difficulty [5], [21]-[24]. And the accumulations of multiple layers with nonlinear activations exacerbates issues related to gradient vanishing or explosion. Hence, we simplify the above graph convolution on the QoS tensor to the following form as:

$$\begin{cases} \mathbf{U}^{(l+1)} = \hat{\mathbf{A}} * \mathbf{S}^{(l)}, \\ \mathbf{S}^{(l+1)} = \hat{\mathbf{A}}^T * \mathbf{U}^{(l)}. \end{cases} \quad (6)$$

On the other hand, it is essential to carefully consider the set to the mixing matrix $\Theta$. The primary focus of this paper is not the intricate design of it. Therefore, in the interest of computational efficiency and overall estimation accuracy, we opt to design $\Theta$ for $\max(0, t-K) \leq i \leq \min(|T|, t+K)$ as:

$$\theta_{t,i} = \frac{1}{\min\left(2K+1, |T|-t+K+1, t+K,\right)}, \quad (7)$$

where $K$ represents the temporal window size for perception, with the other elements in $\Theta$ set to zero. Based on this configuration, the dynamic graph convolution process primarily captures spatiotemporal patterns over more proximate time intervals.

### C. Pooling Features for QoS Estimation

By performing multiple layers of graph convolutions, we obtain the latent feature tensors that encapsulate high-order spatiotemporal information. To further enhance the model's representation learning ability, we aggregate features from different layers using a pooling function as:

$$\begin{cases} \mathbf{U} = P\left(\mathbf{U}^{(0)}, \mathbf{U}^{(1)}, \mathbf{U}^{(2)}, \cdots, \mathbf{U}^{(L)}\right), \\ \mathbf{S} = P\left(\mathbf{S}^{(0)}, \mathbf{S}^{(1)}, \mathbf{S}^{(2)}, \cdots, \mathbf{S}^{(L)}\right), \end{cases} \quad (8)$$

where $P(\cdot)$ can be chosen as *Concatenation*, *Sum*, or *Mean* [5], [24], [28]. Based on the pooling node features, the estimated QoS tensor is achieved as:

$$\hat{\mathbf{Q}} = \mathbf{U} \odot \mathbf{S}^T \quad (9)$$

Following prior studies, this paper employs the Euclidean distance to measure the errors between estimated values and true ones. Consequently, we construct the objective function with an $L_2$ regularization term on known entries of $\mathbf{Q}$ as:

$$\begin{aligned} &\varepsilon\left(\mathbf{U}^{(0)}, \mathbf{S}^{(0)}\right) \\ &= \sum_{q_{u,s,t} \in \Lambda} \left(q_{u,s,t} - \hat{q}_{u,s,t}\right)^2 + \tau\left(\left\|\mathbf{U}^{(0)}\right\|_F^2 + \left\|\mathbf{S}^{(0)}\right\|_F^2\right). \end{aligned} \quad (10)$$

where $\tau$ is the coefficient controlling regularization strength.

## IV. EXPERIMENTS

In this section, we empirically investigate the performance of SCG on two large-scale QoS datasets. First, to validate the QoS estimation capability of SCG, we comprehensively compare it with several state-of-the-art peers. Second, to demonstrate the effectiveness of the proposed spatiotemporal graph convolution, we conduct the ablation studies on SCG. Finally, the sensitivity analysis on important hyperparameters including $L$ and $K$ is presented, providing valuable insights for hyperparameter selection.

### A. General Settings

**Datasets.** Two large-scale widely-adopted dynamic QoS datasets are employed in our study [5]-[11]. They are collected by WSMonitor, which respectively describe the throughput and response time. And for general evaluation, all QoS values have been normalized to the range of 0 to 10. Table I provides the detailed descriptions to them.

TABLE I. DETAILS OF DATASETS

| Dataset | $|U|$ | $|S|$ | $|T|$ | $|\Lambda|$ |
|---|---|---|---|---|
| Throughput | 142 | 4,500 | 64 | 30,287,611 |
| Response Time | 142 | 4,500 | 64 | 30,171,491 |

TABLE II. ACCURACY ON DYNAMIC QoS ESTIMATION

| Method | Throughput | | Response Time | | F-rank |
|---|---|---|---|---|---|
| | RMSE | MAE | RMSE | MAE | |
| *EvolveGCN* | $0.5081_{\pm 3.8E-5}$ | $0.3785_{\pm 8.9E-4}$ | $0.5408_{\pm 1.0E-4}$ | $0.3503_{\pm 1.4E-4}$ | 6.75 |
| *WD-GCN* | $0.4934_{\pm 5.0E-3}$ | $0.3671_{\pm 5.9E-3}$ | $0.4011_{\pm 3.7E-3}$ | $0.2967_{\pm 2.3E-3}$ | 4.50 |
| *TM-GCN* | $0.4145_{\pm 6.3E-6}$ | $0.3185_{\pm 1.4E-4}$ | $0.5063_{\pm 1.1E-5}$ | $0.3814_{\pm 5.3E-4}$ | 4.75 |
| *CTGCN* | $0.3613_{\pm 3.4E-2}$ | $0.2609_{\pm 3.0E-2}$ | $0.3521_{\pm 8.4E-3}$ | $0.2495_{\pm 6.1E-3}$ | 2.00 |
| *PGCN* | $0.4771_{\pm 4.7E-4}$ | $0.3484_{\pm 4.8E-4}$ | $0.3818_{\pm 3.6E-4}$ | $0.2852_{\pm 1.3E-3}$ | 3.50 |
| *MegaCRN* | $0.5009_{\pm 1.2E-3}$ | $0.3745_{\pm 1.2E-3}$ | $0.4083_{\pm 9.7E-4}$ | $0.3070_{\pm 2.5E-3}$ | 5.50 |
| *SCG (Ours)* | $0.2124_{\pm 1.0E-3}$ | $0.1309_{\pm 5.0E-4}$ | $0.2220_{\pm 1.6E-4}$ | $0.1294_{\pm 4.1E-5}$ | 1.00 |

**Evaluation Metrics.** The task of dynamic QoS estimation mainly concerns the estimation accuracy to missing values in a given QoS tensor. Hence, for each tested model, we follow the prior studies to calculate the Root Mean Squared Errors (RMSE) and Mean Absolute Error (MAE) [51]-[62], [76]-80].

**Compared Models.** Six state-of-the-art baselines for dynamic QoS estimation are included in our comparison experiments. They include: EvolveGCN [15], WD-GCN [33], TM-GCN [25], CTGCN [18], PGCN [20], and MegaCRN [13]. These approaches are carefully designed to capture the spatial and temporal information implied in dynamic graph data, but fail to achieve unified modeling of them and suffer from training difficulties owing to attribute dependency and complex message passing rules. Due to the space constraints, the details of them can be found in the original papers.

*B. Comparison with State-of-the-Art Methods*

For a thorough evaluation of the performance of SCG, we have chosen six advanced dynamic graph representation learning methods for comparison. Table II summarizes the RMSE, MAE and the Fridman test results of all models in terms of QoS estimation on the two datasets involved.

We observe that on all datasets, the RMSE and MAE of SCG are significantly lower than those of other comparison models, demonstrating that SCG exhibits superior dynamic QoS estimation performance. For instance, SCG achieves the RMSE at 0.2124 on *Throughput*, which is 58.20%, 56.95%, 48.76%, 41.21%, 55.48%, and 57.60% lower than its peers. The high accuracy of SCG is primarily attributed to the following factors: 1) Effective capture of spatiotemporal patterns enables it to learn richer QoS collaborative signals from spare QoS tensors. 2) Tensor decomposition ensures its stronger adaptability of the latent features. 3) Simplified message propagation architecture and layer pooling mechanism ensure that its parameters are adequately trained.

Besides, the hypothesis that all models involved in our experiments are substantially different is accepted as the significance level is set to 0.01. The high F-ranks of SCG (1.00) and that of other models (6.75, 4.50, 4.75, 2.00, 3.50, and 5.50) summarized in Table II further demonstrates that SCG achieves better performance in dynamic QoS estimation.

*C. Hyperparameter Sensitivity Test*

To investigate how important hyperparameters impact SCG's performance, we conduct experiments to exam the trend of estimation errors with varying values of *L* and *K*, as illustrated in Figs. 2 and 3. From them, we conclude the following findings.

It is expected that increasing the value of *L* allows the SCG model to expand its receptive field on a dynamic graph, thereby capturing higher-order spatiotemporal neighborhood information. The clear downward trend in the error plot depicted in Fig. 2 solidly supports this assertion. For example, on *Response Time*, as *L* is increased from 1 to 7, the RMSE achieved by SCG reduces from 0.2360 to 0.2197 and the MAE reduces from 0.1400 to 0.1278, indicating the error gaps of 6.91% in RMSE and 8.71% in MAE. And the deceleration trend is influenced by the issue of over-smoothing.

Moreover, considering the impact of *K*, its initial increase leads to error reduction. However, subsequently, it may cause the error to increase, as shown in Fig. 3. For instance, on *Throughput*, SCG's RMSE decreases from 0.2197 to 0.2163 as *K* increase from 1 to 4, but increases to 0.2176 as *K* is set to 8. In a similar manner, as the value of *K* varies from 1 to 5, the MAE of SCG initially decreases from 0.1387 to 0.1352, and then worsens, rising to 0.1356. The phenomenon is attributed to the fact that the current and nearby temporal states are more informative, and the distant ones may introduce interference.

*D. Ablation Study*

As discussed in the above analysis, SCG effectively captures both temporal and spatial dependencies, thereby enhancing its QoS estimation accuracy. To validate this, we conduct in-depth ablation studies, whose results are presented in Fig. 4. SCG and

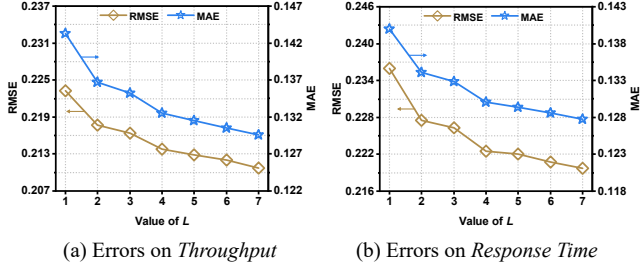

(a) Errors on *Throughput*  (b) Errors on *Response Time*
Figure 2. Estimation errors achieved by SCG as *L* varies.

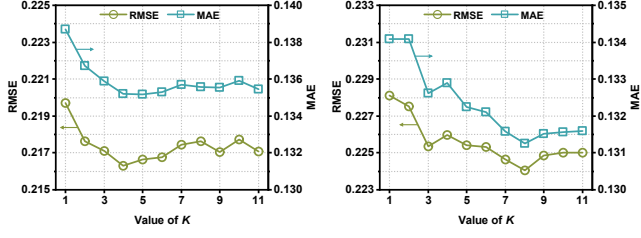

(a) Errors on *Throughput*  (b) Errors on *Response Time*
Figure 3. Estimation errors achieved by SCG as *K* varies.

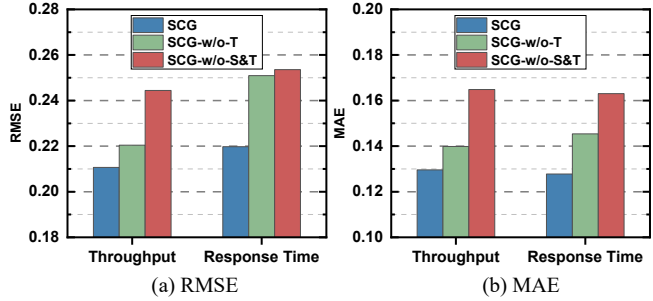

(a) RMSE  (b) MAE
Figure 4. Spatiotemporal characteristic ablation study of SCG.

its two variants including SCG-w/o-T without capture to temporal information, i.e., setting *K* to 0, and SCG-w/o-S&T that removes the capture to both spatial and temporal information involve the process.

As anticipated, the estimation error of SCG is significantly lower than that of its two variants, indicating that its high performance stems from the effective capture of both temporal and spatial patterns. For instance, as plotted in Fig. 4(b), on *Throughput*, SCG achieves the MAE at 0.1296, which is 7.36% lower than that of SCG-w/o-T (0.1399) and 21.41% lower than that of SCG-w/o-T&S (0.1649). Similarly, the MAE on *Response Time* is achieved at 0.1278, the error gap between SCG-w/o-T is 12.10% and the one between SCG-w/o-T&S is 21.64%. Therefore, we conclude from the ablation study that the temporal and spatial information capture modules are significantly beneficial for SCG's dynamic QoS estimation performance.

*E. Summary*

The experimental results and analysis presented above demonstrate the following advantages of SCG in handling dynamic QoS data: 1) ***Simultaneous Handling of temporal and Spatial Patterns:*** SCG effectively captures both temporal and spatial patterns embedded within QoS data, thereby minimizing information loss and achieving comprehensive representation learning. 2) ***Simple yet Powerful Learning Paradigm:*** SCG's message propagation rules are transparent and easily extensible, making it conductive to further improvements. As a result, SCG significantly outperforms its peers in dynamic QoS estimation.

V. CONCLUSIONS

To achieve accurate QoS estimation, this study introduces a novel LFA model, referred to as SCG. Experimental results on two large-scale QoS datasets demonstrate that the proposed SCG outperforms state-of-the-art comparison models in terms of estimation accuracy for missing QoS values. In future work, we intend to further improve SCG's performance by incorporating advanced representation learning methods on HDI data [41]-[50], [63]-[75], [81]-[84]. Regarding the dynamic information propagation mechanism of SCG, further theoretical validation is necessary to show its superior representation learning ability.

REFERENCES

[1] D. Wu, P. Zhang, Y. He, and X. Luo, "A double-space and double-norm ensembled latent factor model for highly accurate web service QoS prediction," *IEEE Trans. on Services Computing*, vol. 16, no. 2, pp. 802-814, 2023.

[2] X. Luo, M. Chen, H. Wu, Z. Liu, H. Yuan, and M. Zhou, "Adjusting learning depth in non-negative latent factorization of tensors for accurately modeling temporal patterns in dynamic QoS data," *IEEE Trans. on Automation Science and Engineering*, vol. 18, no. 4, pp. 2142-2155, 2022.

[3] Y. Yuan, R. Wang, G. Yuan, and X. Luo, "An adaptive divergence-based non-negative latent factor model," *IEEE Trans. on Systems Man Cybernetics: Systems*, vol. 53, no. 10, pp. 6475-6487, 2023.

[4] J. Li, X. Luo, Y. Yuan, and S. Gao, "A nonlinear PID-incorporated adaptive stochastic gradient descent algorithm for latent factor analysis," *IEEE Trans. on Automation Science and Engineering*, 10.1109/TASE.2023.3284819, 2023.

[5] F. Bi, T. He, Y. Xie, and X. Luo, "Two-stream graph convolutional network-incorporated latent feature analysis," *IEEE Trans. on Services Computing*, vol. 16, no. 4, pp. 3027-3042, 2023.

[6] D. Wu, Q. He, X. Luo, M. Shang, Y. He, and G. Wang, "A posterior-neighborhood-regularized latent factor model for highly accurate web service QoS prediction," *IEEE Trans. on Services Computing*, vol. 15, no. 2, pp. 793-805, 2022.

[7] F. Bi, T. He, and X. Luo, "A fast nonnegative autoencoder-based approach to latent feature analysis on high-dimensional and incomplete data," *IEEE Trans. on Services Computing*, DOI: 10.1109/TSC.2023.3319713, 2023.

[8] K. Liu, F. Xue, X. He, D. Guo, and R. Hong, "Joint multi-grained popularity-aware graph convolution collaborative filtering for recommendation," *IEEE Trans. on Computational Social Systems*, vol. 10, no. 1, pp. 72-83, 2023.

[9] D. Wu, X. Luo, M. Shang, Y. He, G. Wang, and X. Wu, "A data-characteristic-aware latent factor model for web services QoS prediction," *IEEE Trans. on Knowledge and Data Engineering*, vol. 34, no. 6, pp. 2525-2538, 2022.

[10] M. Chen, Y. Qiao, R. Wang, and X. Luo, "A generalized Nesterov's accelerated gradient-incorporated non-negative latent-factorization-of-tensors model for efficient representation to dynamic QoS data," *IEEE Trans. on Emerging Topics in Computational Intelligence*, DOI: 10.1109/TETCI.2024.3360338, 2024.

[11] Y. Yuan, X. Luo, M. Shang, and Z. Wang, "A Kalman-filter-incorporated latent factor analysis model for temporally dynamic sparse data," *IEEE Trans. on Cybernetics*, vol. 53, no. 9, pp. 5788-5801, 2023.

[12] E. Tong, W. Niu, and J. Liu, "A missing QoS prediction approach via time-aware collaborative filtering," *IEEE Trans. on Services Computing*, vol. 15, no. 6, pp. 3115-3128, 2022.

[13] R. Jiang, Z. Wang, J. Yong, P. Jeph, Q. Chen, Y. Kobayashi, X. Song, S. Fukushima, and T. Suzumura, "Spatio-temporal meta-graph learning for traffic forecasting," in *Proc. of the 37th AAAI Conf. on Artificial Intelligence*, Washington, USA, 2023, pp. 8078-8086.

[14] Y. Sun, X. Jiang, Y. Hu, F. Duan, K. Guo, B. Wang, J. Gao, and B. Yin, "Dual dynamic spatial-temporal graph convolution network for traffic prediction," *IEEE Trans. on Intelligent Transportation Systems*, vol. 23, no. 12, pp. 23680-23693, 2022.

[15] A. Pareja, G. Domeniconi, J. Chen, T. Ma, T. Suzumura, H. Kanezashi, T. Kaler, T. B. Schardl, and C. E. Leiserson, "EvolveGCN: evolving graph convolutional networks for dynamic graphs," in *Proc. of the 34th AAAI Conf. on Artificial Intelligence*, New York, USA, 2020, pp. 5363-5370.

[16] X. Luo, H. Wu, H. Yuan, and M. Zhou, "Temporal pattern-aware QoS prediction via biased non-negative latent factorization of tensors," *IEEE Trans. on Cybernetics*, vol. 50, no. 5, pp. 1798-1809, 2020.

[17] Y. Yuan, X. Luo, and M. Zhou, "Adaptive divergence-based non-negative latent factor analysis of high-dimensional and incomplete matrices from industrial applications," *IEEE Trans. on Emerging Topics in Computational Intelligence*, DOI: 10.1109/TETCI.2023.3332550, 2023.

[18] J. Liu, C. Xu, C. Yin, W. Wu, and Y. Song, "K-core based temporal graph convolutional network for dynamic graphs," *IEEE Trans. on Knowledge and Data Engineering*, vol. 34, no. 8, pp. 3841-3853, 2022.

[19] A. Cini, I. Marisca, F. M. Bianchi, and C. Alippi, "Scalable spatiotemporal graph neural networks," in *Proc. of the 37th AAAI Conf. on Artificial Intelligence*, Washington, USA, 2023, pp. 7218-7226

[20] Y. Shin, and Y. Yoon, "PGCN: progressive graph convolutional networks for spatial–temporal traffic forecasting," *IEEE Trans. on Intelligent Transportation Systems*, DOI: 10.1109/TITS.2024.3349565, 2024.

[21] W. Cong, S. Zhang, J. Kang, B. Yuan, H. Wu, X. Zhou, H. Tong, and M. Mahdavi, "Do we really need complicated model architectures for temporal networks?" in *Proc. of the 11th Int. Conf. on Learning Representations*, Kigali, Rwanda, 2023.

[22] M. Zhu, X. Wang, C. Shi, H. Ji, and P. Cui, "Interpreting and unifying graph neural networks with an optimization framework," in *Proc. of the 30th ACM Web Conf.*, Ljubljana, Slovenia, 2021, pp. 1215-1226.

[23] T. He, Y. Liu, Y.S. Ong, X. Wu, and X. Luo, "Polarized message-passing in graph neural networks," *Artificial Intelligence*, vol. 331, pp. 104-129, 2024.

[24] X. He, K. Deng, X. Wang, Y. li, Y. Zhang, and M. Wang, "LightGCN: simplifying and powering graph convolution network for recommendation," in *Proc. of the 43rd Int. ACM SIGIR Conf. on Research and Development in Information Retrieva'*, Xi'an, China, 2020, pp. 639-648.

[25] O. A. Malik, S. Ubaru, L. Horesh, M. E. Kilmer, and H. Avron, "Dynamic graph convolutional networks using the tensor M-product," in *Proc. of the 2021 SIAM Int. Conf. on Data Mining*, Virtual Event, 2021, pp. 729-737.

[26] K. Braman, "Third-order tensors as linear operators on a space of matrices," *Linear Algebra and its Applications*, vol. 433, no. 7, pp. 1241-1253, 2010.

[27] X. Luo, Z. Liu, M. Shang, J. Lou, and M. Zhou, "Highly-accurate community detection via pointwise mutual information-incorporated symmetric non-negative matrix factorization," *IEEE Trans. on Network Science and Engineering*, vol. 8, no. 1, pp. 463-476, 2021.

[28] F. Bi, T. He, and X. Luo, "A two-stream light graph convolution network-based latent factor model for accurate cloud service QoS estimation," in *Proc. of the 2022 IEEE Int. Conf. on Data Mining*, Orlando, FL, USA, 2022, pp. 855-860.

[29] T. He, Y. S. Ong, and L. Bai, "Learning conjoint attentions for graph neural nets," in *Proc. of the 34th Annual Conf. on Neural Information Processing Systems*, Virtual Event, 2021, pp. 2641-2653.

[30] K. Xu, W. Hu, J. Leskovec, and S. Jegelka, "How powerful are graph neural networks," in *Proc. of the 7th Int. Conf. on Learning Representations*, New Orleans, USA, 2019.

[31] T. N. Kipf, and M. Welling, "Semi-supervised classification with graph convolutional networks," in *Proc. of the 5th Int. Conf. on Learning Representations*, Toulon, France, 2017.

[32] J. Gilmer, S. S. Schoenholz, P. F. Riley, O. Vinyals, and G. E. Dahl, "Neural message passing for quantum chemistry," in *Proc. of the 34th Int. Conf. on Machine Learning*, Sydney, Australia, 2017, pp. 1263-1272.

[33] F. Manessi, A. Rozza, and M. Manzo, "Dynamic graph convolutional networks," *Pattern Recognition*, vol. 97, 2020.

[34] X. Luo, H. Wu, Z. Wang, J. Wang, and D. Meng, "A novel approach to large-scale dynamically weighted directed network representation," *IEEE Trans.*

*on Pattern Analysis and Machine Intelligence*, vol. 44, no. 12, pp. 9756-9773, 2022.

[35] T. Chen, S. Li, Y. Qiao, and X. Luo, "A robust and efficient ensemble of diversified evolutionary computing algorithms for accurate robot calibration," *IEEE Trans. on Instrumentation and Measurement*, vol. 73, pp. 1-14, 2024.

[36] Y. Zhong, L. Jin, M. Shang, and X. Luo, "Momentum-incorporated symmetric non-negative latent factor models," *IEEE Trans. on Big Data*, vol. 8, no. 4, pp. 1096-1106, 2022.

[37] F. Bi, X. Luo, B. Shen, H. Dong, and Z. Wang, "Proximal alternating-direction-method-of-multipliers-incorporated nonnegative latent factor analysis," *IEEE/CAA Journal of Automatica Sinica*, vol. 10, no. 6, pp. 1388-1406, 2023.

[38] X. Luo, Y. Yuan, M. Zhou, Z. Liu, and M. Shang, "Non-negative latent factor model based on β-divergence for recommender systems," *IEEE Trans. on Systems Man Cybernetics: Systems*, vol. 51, no. 8, pp. 4612-4623, 2021.

[39] X. Glorot, and Y. Bengio, "Understanding the difficulty of training deep feedforward neural networks," in *Proc. of the 13th Int. Conf. on Artificial Intelligence and Statistics,* Sardinia, Italy, 2010, pp. 249-256.

[40] D. P. Kingma, and J. Ba, "Adam: a method for stochastic optimization," *arXiv preprint arXiv:1412.6980*, 2014.

[41] X. Luo, W. Qin, A. Dong, K. Sedraoui, and M. C. Zhou, "Efficient and High-quality Recommendations via Momentum-incorporated Parallel Stochastic Gradient Descent-based Learning," *IEEE/CAA Journal of Automatica Sinica*, vol. 8, no. 2, pp. 402-411, 2021.

[42] X. Luo, Y. Yuan, S. L. Chen, N. Y. Zeng, and Z. D. Wang, "Position-Transitional Particle Swarm Optimization-Incorporated Latent Factor Analysis," IEEE Trans. on Knowledge and Data Engineering, DOI: 10.1109/TKDE.2020.3033324, 2020.

[43] X. Luo, M. C. Zhou, Y. N. Xia, and Q. S. Zhu, "An efficient non-negative matrix-factorization-based approach to collaborative filtering for recommender systems," *IEEE Trans. on IndustrialInformatics*, vol. 10, no. 2,pp. 1273‑1284, 2014.

[44] X. Luo, M. C. Zhou, S. Li, and M. S. Shang, "An inherently non‑negative latent factor model for high‑dimensional and sparse matrices from industrial applications," *IEEE Trans. on IndustrialInformatics*, vol. 14, no. 5, pp. 2011–2022, 2018.

[45] X. Luo, Z. D. Wang, and M. S. Shang, "An Instance-frequency-weighted Regularization Scheme for Non-negative Latent Factor Analysis on High Dimensional and Sparse Data," *IEEE Trans. on System Man Cybernetics: Systems*, vol. 51, no. 6, pp. 3522-3532, 2021.

[46] L. Hu, X. H. Yuan, X. Liu, S. W. Xiong, and X. Luo, "Efficiently Detecting Protein Complexes from Protein Interaction Networks via Alternating Direction Method of Multipliers," *IEEE/ACM Trans. on Computational Biology and Bioinformatics*, vol. 16, no. 6, pp. 1922-1935, 2019.

[47] L. Hu, P. W. Hu, X. Y. Yuan, X. Luo, and Z. H. You, "Incorporating the Coevolving Information of Substrates in Predicting HIV-1 Protease Cleavage Sites," *IEEE/ACM Trans. on Computational Biology and Bioinformatics*, vol. 17, no. 6, pp. 2017-2028, 2020.

[48] X. Luo, Z. G. Liu, S. Li, M. S. Shang, and Z. D. Wang, "A Fast Non-negative Latent Factor Model based on Generalized Momentum Method," *IEEE Trans. on System, Man, and Cybernetics: Systems*, vol. 51, no. 1, pp. 610-620, 2021.

[49] Z. G. Liu, X. Luo, and Z. D. Wang, "Convergence Analysis of Single Latent Factor-Dependent, Nonnegative, and Multiplicative Update-Based Nonnegative Latent Factor Models," *IEEE Trans. on Neural Networks and Learning Systems*, vol. 32, no. 4, pp. 1737-1749, 2021.

[50] Y. Yuan, Q. He, X. Luo, and M. S. Shang, "A multilayered-and-randomized latent factor model for high-dimensional and sparse matrices," *IEEE Trans. on Big Data*, DOI: 10.1109/TBDATA.2020.2988778, 2020.

[51] X. Luo, Z. Li, W. Yue and S. Li, "A Calibrator Fuzzy Ensemble for Highly-Accurate Robot Arm Calibration," *IEEE Transactions on Neural Networks and Learning Systems*, DOI: 10.1109/TNNLS.2024.3354080.

[52] J. Chen, Y. Yuan and X. Luo, "SDGNN: Symmetry-Preserving Dual-Stream Graph Neural Networks," *IEEE/CAA Journal of Automatica Sinica*, vol. 11, no. 7, pp. 1717-1719, 2024.

[53] S. Sedhain, A. K. Menon, S. Sanner, and L. Xie, "Autorec: autoencoders meet collaborative filtering," in *Proc. of the Int. Conf. on World Wide Web*, pp. 111-112, 2015.

[54] Q. Yao, X. Chen, J. T. Kwok, Y. Li, and C. J. Hsieh, "Efficient neural interaction function search for collaborative filtering," in *Proc. of the Web Conference*, pp. 1660-1670, 2020.

[55] K. L. Elmore and M. B. Richman, "Euclidean Distance as a Similarity Metric for Principal Component Analysis," *Monthly Weather Review*, vol. 129, no. 3, pp. 540-549, 2010.

[56] W. Qin and X. Luo, "Asynchronous Parallel Fuzzy Stochastic Gradient Descent for High-Dimensional Incomplete Data Representation," *IEEE Transactions on Fuzzy Systems*, vol. 32, no. 2, pp. 445-459, Feb. 2024.

[57] W. Qin, X. Luo and M. Zhou, "Adaptively-Accelerated Parallel Stochastic Gradient Descent for High-Dimensional and Incomplete Data Representation Learning," *IEEE Transactions on Big Data*, vol. 10, no. 1, pp. 92-107,. 2024.

[58] X. Luo, H. Wu and Z. Li, "Neulft: A Novel Approach to Nonlinear Canonical Polyadic Decomposition on High-Dimensional Incomplete Tensors," in *IEEE Transactions on Knowledge and Data Engineering*, vol. 35, no. 6, pp. 6148-6166, 2023.

[59] P. Massa, and P. Avesani, "Trust-aware recommender systems," in *Proc. of the 1st ACM Conf. on Recommender Systems*, pp. 17-24, 2007.

[60] S. Zhang, L. Yao, and X. Xu, "Autosvd++ an efficient hybrid collaborative filtering model via contractive auto-encoders," in *Proc. of the 40th International ACM SIGIR conference on Research and Development in Information Retrieval*, pp. 957-960, 2017.

[61] Y. Cao, W. Shi, L. Sun, and X. Fu, "Channel State Information Based Ranging for Underwater Acoustic Sensor Networks," *IEEE Trans. on Wireless Communications*, DOI: 10.1109/TWC.2020.3032589, 2020.

[62] X. Luo, Y. Zhou, Z. Liu and M. Zhou, "Fast and Accurate Non-Negative Latent Factor Analysis of High-Dimensional and Sparse Matrices in Recommender Systems," *IEEE Transactions on Knowledge and Data Engineering*, vol. 35, no. 4, pp. 3897-3911, 2023.

[63] D. Wu, X. Luo, Y. He and M. Zhou, "A Prediction-Sampling-Based Multilayer-Structured Latent Factor Model for Accurate Representation to High-Dimensional and Sparse Data," *IEEE Transactions on Neural Networks and Learning Systems*, vol. 35, no. 3, pp. 3845-3858, 2024.

[64] H. Wu, X. Luo, M. Zhou, M. J. Rawa, K. Sedraoui and A. Albeshri, "A PID-incorporated Latent Factorization of Tensors Approach to Dynamically Weighted Directed Network Analysis," *IEEE/CAA Journal of Automatica Sinica*, vol. 9, no. 3, pp. 533-546, 2022.

[65] L. Hu, S. Yang, X. Luo and M. Zhou, "An Algorithm of Inductively Identifying Clusters From Attributed Graphs," *IEEE Transactions on Big Data*, vol. 8, no. 2, pp. 523-534, 2022.

[66] X. Luo, M. Chen, H. Wu, Z. Liu, H. Yuan and M. Zhou, "Adjusting Learning Depth in Nonnegative Latent Factorization of Tensors for Accurately Modeling Temporal Patterns in Dynamic QoS Data," *IEEE Transactions on Automation Science and Engineering*, vol. 18, no. 4, pp. 2142-2155, 2021.

[67] L. Hu, J. Zhang, X. Pan, X. Luo and H. Yuan, "An Effective Link-Based Clustering Algorithm for Detecting Overlapping Protein Complexes in Protein-Protein Interaction Networks," *IEEE Transactions on Network Science and Engineering*, vol. 8, no. 4, pp. 3275-3289, 2021.

[68] X. Luo, Y. Zhong, Z. Wang and M. Li, "An Alternating-Direction-Method of Multipliers-Incorporated Approach to Symmetric Non-Negative Latent Factor Analysis," *IEEE Transactions on Neural Networks and Learning Systems*, vol. 34, no. 8, pp. 4826-4840, 2023.

[69] W. Li, X. Luo, H. Yuan and M. Zhou, "A Momentum-Accelerated Hessian-Vector-Based Latent Factor Analysis Model," in IEEE Transactions on Services Computing, vol. 16, no. 2, pp. 830-844, 2023.

[70] H. Li, K. Li, J. An, W. Zheng, and K. Li, "An efficient manifold regularized sparse non-negative matrix factorization model for large-scale recommender systems on GPUs," *Information Sciences*, vol. 496, pp. 464-484, 2019.

[71] H. Wu, Y. Xia and X. Luo, "Proportional-Integral-Derivative-Incorporated Latent Factorization of Tensors for Large-Scale Dynamic Network Analysis," in *2021 China Automation Congress*, Beijing, China, 2021, pp. 2980-2984.

[72] S. Zhang, L. Yao, B. Wu, X. Xu, X. Zhang, and L. Zhu, "Unraveling Metric Vector Spaces With Factorization for Recommendation," *IEEE Trans. on Industrial Informatics*, vol. 16, no. 2, pp. 732-742, 2020.

[73] J. Li, J. M. Bioucas-Dias, A. Plaza, and L. Liu, "Robust collaborative nonnegative matrix factorization for hyperspectral unmixing," *IEEE Trans. on Geoscience and Remote Sensing*, vol. 54, no. 10, pp. 6076–6090, 2016.

[74] Z. Lin and H. Wu, "Dynamical Representation Learning for Ethereum Transaction Network via Non-negative Adaptive Latent Factorization of Tensors," in *2021 International Conference on Cyber-Physical Social Intelligence*, Beijing, China, 2021, pp. 1-6.

[75] F. Bi, and D. Wu, "A Proximal Alternating-direction-method-of-multipliers-based Nonnegative Latent Factor Model," in *2021 IEEE International Conference on Big Knowledge*, Auckland, New Zealand, 2021, pp. 353-360.

[76] P. Paatero, and U. Tapper, "Positive matrix factorization: a non-negative factor model with optimal utilization of error estimates of data values," *Environmetrics*, vol. 5, no. 2, pp. 111-126, 1994.

[77] N. Guan, D. Tao, Z. Luo, and B. Yuan, "Nenmf: An optimal gradient method for nonnegative matrix factorization," *IEEE Trans. on SiADNLF Processing*, vol. 60, no. 6, pp. 2882–2898, 2012.

[78] S. Zhang, W. Wang, J. Ford, and F. Makedon, "Learning from Incomplete Ratings Using Non-negative Matrix Factorization," in *Proc. of the SIAM Int. Conf. on Data Mining*, pp. 549-553, 2006.

[79] N. Zeng, X. Li, P. Wu, H. Li and X. Luo, "A Novel Tensor Decomposition-Based Efficient Detector for Low-Altitude Aerial Objects With Knowledge Distillation Scheme," *IEEE/CAA Journal of Automatica Sinica*, vol. 11, no. 2, pp. 487-501, February 2024, DOI: 10.1109/JAS.2023.

[80] X. Luo, L. Wang, P. Hu and L. Hu, "Predicting Protein-Protein Interactions Using Sequence and Network Information via Variational Graph Autoencoder," *IEEE/ACM Transactions on Computational Biology and Bioinformatics*, vol. 20, no. 5, pp. 3182-3194, 2023.

[81] H. Zhou, T. He, Y. -S. Ong, G. Cong and Q. Chen, "Differentiable Clustering for Graph Attention," *IEEE Transactions on Knowledge and Data Engineering*, vol. 36, no. 8, pp. 3751-3764, 2024.

[82] T. He and K. C. C. Chan, "Discovering Fuzzy Structural Patterns for Graph Analytics," *IEEE Transactions on Fuzzy Systems*, vol. 26, no. 5, pp. 2785-2796, 2018.

[83] T. He and K. C. C. Chan, "MISAGA: An Algorithm for Mining Interesting Subgraphs in Attributed Graphs," *IEEE Transactions on Cybernetics*, vol. 48, no. 5, pp. 1369-1382.

[84] Y. Xie, Y. Liang, M. Gong, A. K. Qin, Y. -S. Ong and T. He, "Semisupervised Graph Neural Networks for Graph Classification," *IEEE Transactions on Cybernetics*, vol. 53, no. 10, pp. 6222-6235, 2023.